\documentclass[journal]{IEEEtran}
\usepackage{multicol}
\usepackage{subfig}
\usepackage{stfloats}
\usepackage{hyperref}
\usepackage{url}
\usepackage{booktabs} % For formal tables
\usepackage{multirow}
\usepackage{tabularx}
\usepackage {graphicx}
\usepackage{array}
\usepackage{amssymb}
\usepackage{dsfont}

\ifCLASSINFOpdf
\else
\fi
\begin{document}
\title{Neural method for Explicit Mapping of Quasi-curvature Locally Linear Embedding in image retrieval}

\author{Shenglan~Liu, Jun~Wu, Lin~Feng, Feilong~Wang
\thanks{
Shenglan Liu  and Lin Feng  are with Faculty of Electronic Information and Electrical Engineering, Dalian University of Technology, Dalian, Liaoning, 116024 China. Jun Wu and Feilong Wang is with the School of Innovation and Entrepreneurship, Dalian University of Technology, Dalian, Liaoning, 116024 China. e-mail: ( \{liusl, fenglin, wangfeilong\}@dlut.edu.cn).}
}


\markboth{Journal of \LaTeX\ Class Files}%
{Shell \MakeLowercase{\textit{et al.}}: Bare Demo of IEEEtran.cls for IEEE Journals}



\maketitle

\begin{abstract}
This paper proposed a new explicit nonlinear dimensionality reduction using neural networks for image retrieval tasks. We first proposed a Quasi-curvature Locally Linear Embedding (QLLE) for training set. QLLE guarantees the linear criterion in neighborhood of each sample. Then, a neural method (NM) is proposed for out-of-sample problem. Combining  QLLE and NM, we provide a explicit nonlinear dimensionality reduction approach for efficient image retrieval. The experimental results in three benchmark datasets illustrate that our method can get better performance than other state-of-the-art out-of-sample methods.
\end{abstract}


% \fi
\IEEEpeerreviewmaketitle

\section{Introduction}
\IEEEPARstart{R}{ecently}  manifold learning and dimensionality reduction have attracting more and more attentions since high-dimensional data have brought lots of computational problems in many applications\cite{4, 7}. In computer vision, visual images embedded in high-dimensional feature space actually have relatively low-dimensional representations which preserve the intrinsic components in images. So many superior dimensionality reduction algorithms have been applied for low-dimensional image representations to accelerate the subsequent retrieval process since the feature dimension affects directly the time complexity in similarity-based ranking.

Many classical dimensional dimensionality algorithms have been proposed in recent years. Traditional linear techniques such as principal component analysis (PCA)\cite{3} have shown great efficiency and effectiveness in computer vision. In addition, nonlinear algorithms are presented to deal with various data with more complicated structure, such as Laplacian Eigenmaps (LE)\cite{1}, Locally Linear Embedding (LLE)\cite{8}, Local Tangent Space Alignment (LTSA)\cite{11}, etc. The latter ones assumed that the underlying manifold can be approximated in the form of adjacent graph. And these algorithms tend to be transformed into the problems of eigen-decomposition of spectral matrix. But it is very time-consuming for large amount of samples, which limits the performance in many vision tasks.

There are little research works about the out-of-sample problems for low-dimensional feature representation. Vladymyrov et.al\cite{9} presented locally linear landmarks (LLL) algorithm for large-scale manifold learning,
which solve for a smaller graph defined on selected landmarks and then apply
the Nystrom formula to estimate the eigenvectors over all points. For large-scale
manifold learning, it is not a reasonable choice for classical spectral algorithms
to solve the expensive eigen-decomposition.

In this paper, we propose a novel learning framework for large-scale dimensionality reduction, which learns the approximating manifold mapping from small landmarks. A small subset of original data are chosen to find the low-dimensional embedding, and then the explicit mapping function from original points to low-dimensional coordinates can be learnt using these small subset. Therefore, the main contributions in this paper are:
(1) A quasi-curvature index is proposed to measure approximately the curvatures of local neighborhoods, and then Quasi-curvature Locally Linear Embedding (QLLE) is presented for non-linear manifold learning.
(2) A novel large-scale manifold learning framework is proposed based on the idea that the mapping function from original points to low-dimensional coordinates can be learned explicitly using small landmarks.

\section{Quasi-curvature Locally Linear embedding and Out-of-sample by Neural Method}

In this section, we propose QLLE in subsection 2.1, which has superior efficiency and effectiveness in nonlinear
dimensionality reduction for small data sampled from some underlying
manifold. Then, landmarks chosen from original data can be utilized to get
low-dimensional coordinates using QLLE; and out-of-sample problem can be solved by Neural Method (NM) in subsection 2.2 and 2.3.

\subsection{Quasi-curvature Locally Linear Embedding}

LLE considers that local linear should be guaranteed in neighborhood of each sample. However, local linear is a very strong assumption for real word data. From this view of point, we involve curvature to evaluate the linear condition of neighborhood of each sample. Adaptive neighbor selection is also proposed in QLLE. The detail of QLLE can be concluded in the following Algorithm 1.

\begin{table*}[htbp]
\begin{center}
\begin{tabular}
{p{35pt}p{350pt}}%
\hline
Alg. 1 & Quasi-curvature Locally Linear Embedding \\
\hline
Input: & The original dataset $\hat {X} = \left[ {x_1 ,x_2 , \cdots ,x_P } \right] \in {\mathds{R}}^{D\times P}$ , neighborhood parameter $k$, curvature threshold $\eta$ and low-dimensional dimensions $d$.\\
Output: & low-dimensional coordinates $\hat {Y} = \left[ {y_1 ,y_2 , \cdots ,y_P } \right] \in {\mathds{R}}^{d\times P}$
 \\
\hline
Step 1& Initialization KNN of $\hat {X}$: $NI = \left\{ {\left. {N_k \left( {x_i } \right)} \right|\mathop {\min }\limits_{j = \{i_1 , \cdots ,i_k \}} \left\| {x_i - x_j } \right\|,i = 1, \cdots ,P} \right\}$ ;
 \\

Step 2&
Update $NI_i $ and local construction weights $W = \left[ {w_1 , \cdots w_P
} \right] \in {\mathds{R}}^{P\times P}$ .

\textbf{\textit{for}} $i$=1:$P$

\begin{enumerate}[itemindent=0em]

\item[] For each $NI_i $, computing curvature parameter $c_i $ by $c_i =
{\sum\nolimits_{j = 1}^k {\sqrt {\left\| {Q_i \tilde {x}_{ij}^e }
\right\|^2} } } \mathord{\left/ {\vphantom {{\sum\nolimits_{j = 1}^k {\sqrt
{\left\| {Q_i \tilde {x}_{ij}^e } \right\|^2} } } k}} \right.
\kern-\nulldelimiterspace} k$, where $\tilde {x}_{ij}^e = {\left( {x_{ij} -
\bar {x}_i } \right)} \mathord{\left/ {\vphantom {{\left( {x_{ij} - \bar
{x}_i } \right)} {\left\| {x_{ij} - \bar {x}_i } \right\|}}} \right.
\kern-\nulldelimiterspace} {\left\| {x_{ij} - \bar {x}_i } \right\|}$, $\bar
{x}_i = {\sum\nolimits_{j = 1}^k {x_{ij} } } \mathord{\left/ {\vphantom
{{\sum\nolimits_{j = 1}^k {x_{ij} } } k}} \right. \kern-\nulldelimiterspace}
k$,$x_{ij} \in NI_i $; $Q_i $ is a basis matrix which is calculated by PCA of
$NI_i $, and $c_{ij} = {\sqrt {\left\| {Q_i \tilde {x}_{ij}^e } \right\|^2} }
\mathord{\left/ {\vphantom {{\sqrt {\left\| {Q_i \tilde {x}_{ij}^e }
\right\|^2} } k}} \right. \kern-\nulldelimiterspace} k$.
\end{enumerate}
\noindent
{\kern 50pt}\textbf{\emph{if}} $c_{ij} > \eta $

\[
NI_i = NI_i - \{x_{ij} \},k_i = \left| {NI_i } \right|;
\]

\noindent
{\kern 50pt}\textbf{\emph{end}}

\begin{enumerate}[itemindent=0em]
\item[] Computing $W$ by the follow minimization: $\mathop {\min }\limits_{w_i }
\sum\nolimits_{i = 1}^P {\left\| {x_i - \sum\nolimits_{j = 1}^{k_i } {w_{ij}
x_{ij} } } \right\|} $;
\end{enumerate}
\textbf{\textit{end}}
 \\

Step 3 & Computing low-dimensional embedding $\hat{Y}$ by the following optimization
problem: $\mathop {\min }\limits_{\hat {Y}} \sum\nolimits_{i = 1}^P {c_i \left\| {y_i -
\sum\nolimits_{j = 1}^{k_i } {w_{ij} y_{ij} } } \right\|} ^2$.
  \\
\hline
\end{tabular}
\end{center}
\end{table*}

QLLE considers the local linear configurations among the
whole data, and thus the large nearest neighbor matrix brings a great amount
of calculation and memory in the subsequent process. These problems may
limit the applicability of QLLE in analyzing the nonlinear embedding for
large-scale dataset. In this section, based on the idea in QLLE, we further
propose a modified nonlinear learning algorithm called Neural Method (NM). The basic idea of this algorithm is to learn the
projection relationship between selective small-scale data and the
corresponding low-dimensional coordinates based on QLLE. And then the
learning function can be used to learn the rest of data directly.

\subsection{The selected landmarks}

Suppose data $\hat{X} = [\hat{x}_1, \hat{x}_2, \cdots, \hat{x}_P]\in {\mathds{R}}^{d\times P}$ contains $P$ distinct samples chosen from original data $X$ with the dimensionality $D$. Based on the analysis in QLLE,
the time complexity is about $O\left( {P^2} \right)$, which is acceptable for
reconstructing global low-dimensional coordinates of landmarks when the
number of chosen landmarks is much smaller than that of original data in
large-scale datasets, $P\ll N$. Besides, the number of landmarks should slightly
smaller than or equal to that of original data in small datasets, $P\leq N$,
because QLLE is difficult to reconstruct neighborhood weights when landmarks
are too much few or distributed non-uniformly (or too sparsely). After
computing in QLLE algorithm, the low-dimensional coordinates $\hat {Y} = \left[ {\hat {y}_1 ,\hat {y}_2 , \cdots ,\hat {y}_P } \right] \in {\mathds{R}}^{d\times P}$ can be obtained.

The selection of landmarks in this step has some options. One common method
is to select landmarks randomly without any strategy. It is computationally
efficient and robust for our algorithm though the selected landmarks might
have randomness. Otherwise, we can adopt some clustering algorithms\cite{10} to choose the cluster centers as the landmarks. With these
algorithms, the landmarks are more likely to approximate the true manifold
structure of all the samples, but the computational complexity is also
increasing. So we choose to randomly select the landmarks from original
data. In addition, the parameter $P$ also needs to be determined. Obviously,
the more the number of landmarks is, the more approximate they are to true
manifold structure. Considering both the time complexity of landmarks in
QLLE and the approximation of these landmarks to true manifold structure, an
appropriate parameter $P$ needs to be considered carefully in the experiments.

\subsection{Extreme learning machine for landmarks}

Given the landmarks $\hat{X}$ and their low-dimensional coordinates $\hat
{Y}$, the mapping function between them can be learned with Extreme Learning
Machine (ELM)\cite{2}. The idea of ELM is that the input weights and hidden layer
biases can be randomly generated so that only the output weights need to be
determined. So the learning problem can be transformed into a simple linear
system in which the output weights can be analytically determined through a
generalized inverse operation of the hidden layer weight matrices.

Supposed that the standard ELM has $\tilde{N}$ hidden nodes, and we choose
sigmoid function as the activation function $g\left( x \right)$ in ELM. The
learning problem in ELM can be modeled as Eq.(1):

\begin{equation}
\label{eq7}
\sum\limits_{i = 1}^{\tilde{N}} {\beta _{i}g(a_{i}\hat {x}_{j}+b_{j}) = \hat{y}_{j} }
\end{equation}

where $a_i = \left[ {a_{i1} ,a_{i2} ,\ldots ,a_{in} } \right]^T$ represents
the weight vectors connecting an i th hidden node to the input nodes and
$\beta _i = \left[ {\beta _{i1} ,\beta _{i2} ,\ldots ,\beta _{in} }
\right]^T$ represents the weight vectors connecting the $i-th$ hidden node to
the output nodes. $b_i $ is the threshold for the $i-th$ hidden node and $a_i
\cdot \hat {x}_i $ is the inner product of $a_i$ and $\hat {x}_i $. And
then the above equation can be further rewritten compactly as:

\begin{equation}
\label{eq8}
H\beta = \hat {Y}
\end{equation}

where

$H = \left[ {{\begin{array}{*{20}c}
 {g(a_1 ,\hat {x}_1 ,b_1 )} \hfill & \cdots \hfill & {g(a_{\tilde {N}} ,\hat
{x}_1 ,b_{\tilde {N}} )} \hfill \\
 \vdots \hfill & \cdots \hfill & \vdots \hfill \\
 {g(a_1 ,\hat {x}_P ,b_1 )} \hfill & \cdots \hfill & {g(a_{\tilde {N}} ,\hat
{x}_P ,b_{\tilde {N}} )} \hfill \\
\end{array} }} \right]_{P\times \tilde {N}} \quad ,
\quad$\\$
\beta = \left[ {{\begin{array}{*{20}c}
 {\beta _1^T } \hfill \\
 \vdots \ \\
 {\beta _{\tilde {N}}^T } \hfill \\
\end{array} }} \right]_{\tilde {N}\times d} $ and $\hat {Y} = \left[
{{\begin{array}{*{20}c}
 {\hat {y}_1^T } \hfill \\
 \vdots \hfill \\
 {\hat {y}_P^T } \hfill \\
\end{array} }} \right]_{P\times d} $

In Eq.(2), $H$ is called the hidden layer output matrix of ELM and the $i-th$ column of $H$
is the output of $i-th$ hidden node corresponding to the inputs $\hat {x}_1
,\hat {x}_2 , \cdots ,\hat {x}_P $. Huang et al. has proved that
the hidden layer parameters can be randomly assigned to avoid constantly
being adjusted in traditional paradigm if the activation function $g$is
infinitely differentiable in any interval. So for the fixed network
parameters, the learning of ELM is simply equal to find a least-square
solution of the linear system:

\begin{equation}
\label{eq9}
\mathop {\min }\limits_\beta \left\| {H\beta - \hat {Y}} \right\|
\end{equation}

By solving (3), the least-square solution of the above linear system can be written as
$\beta = H^\dag \hat {Y}$ where $H^\dag $ is the Moore-Penrose generalized
inverse of $H$. Finally, we obtain the explicit mapping function from
original space to low-dimensional coordinates:

\begin{equation}
\label{eq10}
f\left( x \right) = \sum\limits_{i = 1}^{\tilde {N}} {\beta _i g\left( {a_i
\cdot x + b_i } \right)}
\end{equation}

In Eq.(4), we can see that ELM has superior learning efficiency because the only parameters needed to
be determined are the weights of output layers. And thus we choose ELM to
learn approximately the explicit expression of non-linear mapping function
in QLLE using only landmarks.

We can use these landmarks and ELM approach
to find out the learning function to reconstruct the rest data in
low-dimensional coordinates. Therefore, there are three main steps in this
algorithm:\\
1) find the low-dimensional coordinates of selected landmarks;\\
2) give the explicit mapping function with landmarks;\\
3) utilize this function to find the low-dimensional coordinates of the rest data.\\

\subsection{Dimensionality reduction}

The fundamental objection in manifold learning algorithms is to find out a
mapping function which maps the original data to low-dimensional coordinates
meanwhile preserves the manifold structures among data points. When
confronted with large-scale datasets, data points are densely distributed in
high-dimensional space. Due to the high density in data, small points can be
selected as landmarks to approximate the manifold structure in original
space. Under this circumstance, the mapping function learned from these
landmarks can also utilize for dimensionality reduction for other data
points. But the nonlinear QLLE algorithm has no explicit expression for the
mapping function. ELM is used to learn approximately explicit expression of
non-linear mapping function in QLLE.

Once the explicit mapping function $f\left( x \right)$ is determined, the
low-dimensional coordinates of all the samples can be learned directly.
Furthermore, this function can also be used to learn the out-of-samples
without further modifying the learning model.

\section{Experimental results and analysis}
\subsection{Datasets}
Extensive experiments are conducted to evaluate the effectiveness of our proposed out-of-sample dimensionality reduction. Three benchmark image datasets (Corel-1K, Corel-10K and Cifar10) are adopted in this section. The details (Image Size (IS), Number of Categories (NC), Number of Each Categories (NEC), Total Images (TI)) of the three datasets are listed in Table \ref{tab1}.

\begin{center}
\begin{table}[htbp]
\caption{Attributes of experimental dataset}
\begin{tabular}
{cccccc}
\hline
Database&
IS&
NC&
NEC&
TI&
$k$ \\
\hline
Corel-1K&
384$\times $256&
10&
100&
1000&
10 \\

Corel-10K&
Vary&
100&
100&
10000&
10 \\

Cifar-10&
32$\times $32&
10&
6000&
60000&
10 \\
\hline
\end{tabular}
\label{tab1}
\end{table}
\end{center}

\subsection{ Experimental results }
	In this subsection, we utilize PUD \cite{liu2016perceptual} and GIST \cite{oliva2001modeling} descriptors to evaluate LLE* \footnote{LLE* indicates QLLE for convenience.} and corresponding out-of-sample approaches. The PUD (used in Corel-1K and Corel-10K) is a 280 dimensions manifold-based image descriptor and GIST (Cifar-10) is also a perceptive descriptor with 512 dimensions. In our ELM-based out-of-sample method, we set 1000 hidden nodes using sigmoid kernel function. LLE* is executed by using $k=8$ and $d={10,20,...,100}$.  The mean precision (Mea.P.) and maximum precision (Max.P.) of three datasets using different $d$ are listed in Table \ref{tab2}. Table \ref{tab2} contains precision comparison of NM-LLE*, LLL-LLE*, LLE*, PCA and original feature. We set 20, 12 and 500 returns in Corel-1K, Corel-10K and Cifar-10 respectively. NM-LLE* and LLL-LLE* out-of-sample methods utilize 600, 2000 and 3000 landmarks for LLE* dimensionality reduction in Corel-1K, Corel-10K and Cifar-10 respectively. Mean time (Mea.T.) consuming of NM-LLE*, LLL-LLE*, LLE*, PCA and original feature are listed in Table \ref{tab3}. We claim the time consuming of NM-LLE*, LLL-LLE*, LLE* and PCA including dimensionality reduction time, which is fair to original feature experiments.

\begin{center}
\begin{table*}[htbp]
\caption{The precision of four dimensionality reduction methods in three datasets (\%)}
\centerline{
\begin{tabular}
{cccccccccc}
\hline
\multirow{3}{*}{Datasets}&
\multicolumn{8}{c}{Methods} &
\multirow{3}{*}{Original Feature} \\
\cline{2-9}
& \multicolumn{2}{c}{NM-LLE*}
& \multicolumn{2}{c}{LLL-LLE*}
& \multicolumn{2}{c}{LLE*}
& \multicolumn{2}{c}{PCA}
\\
\cline{2-9}
& Mea.P.&
Max.P.&
Mea.P.&
Max.P.&
Mea.P.&
Max.P.&
Mea.P.&
Max.P.\\
\hline
Corel-1K&
74.21&
80.68&
71.22&
78.43&
69.21&
79.36&
72.23&
77.12&
76.67\\
Corel-10K&
45.60&
53.85&
38.44&
43.60&
39.32&
47.27&
44.63&
50.47&
50.24\\
Cifar-10&
28.75&
31.22&
24.81&
26.12&
-&
-&
28.41&
30.55&
30.07\\
\hline
\end{tabular}}
\label{tab2}
\end{table*}
\end{center}

\begin{figure}[htbp]
\centerline{\includegraphics[width=3.2 in]{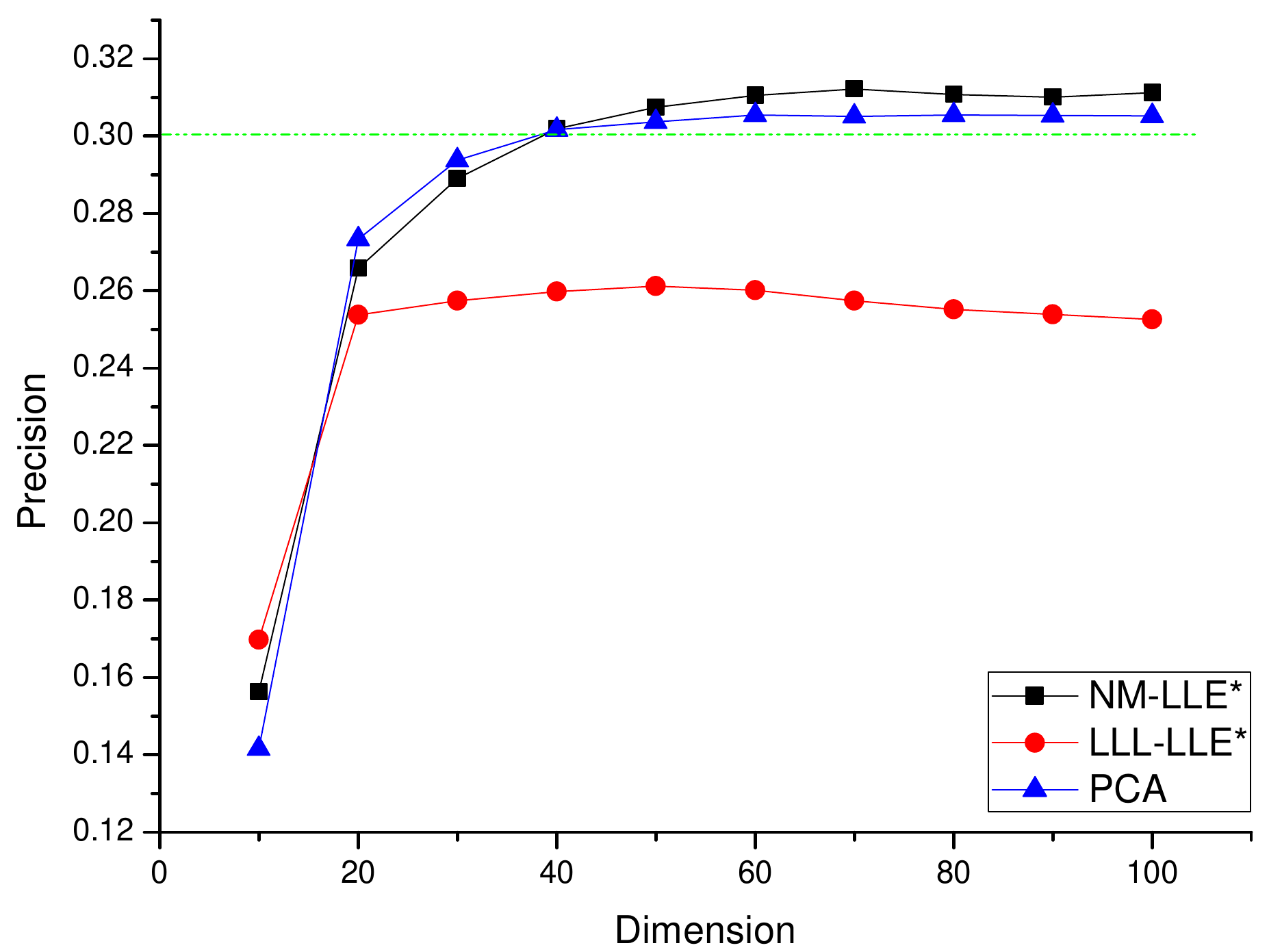}}
\label{fig3}
\caption{The precision of the four methods and original feature in Cifar-10}
\end{figure}

\begin{figure}[htbp]
\centerline{\includegraphics[width=3.2 in]{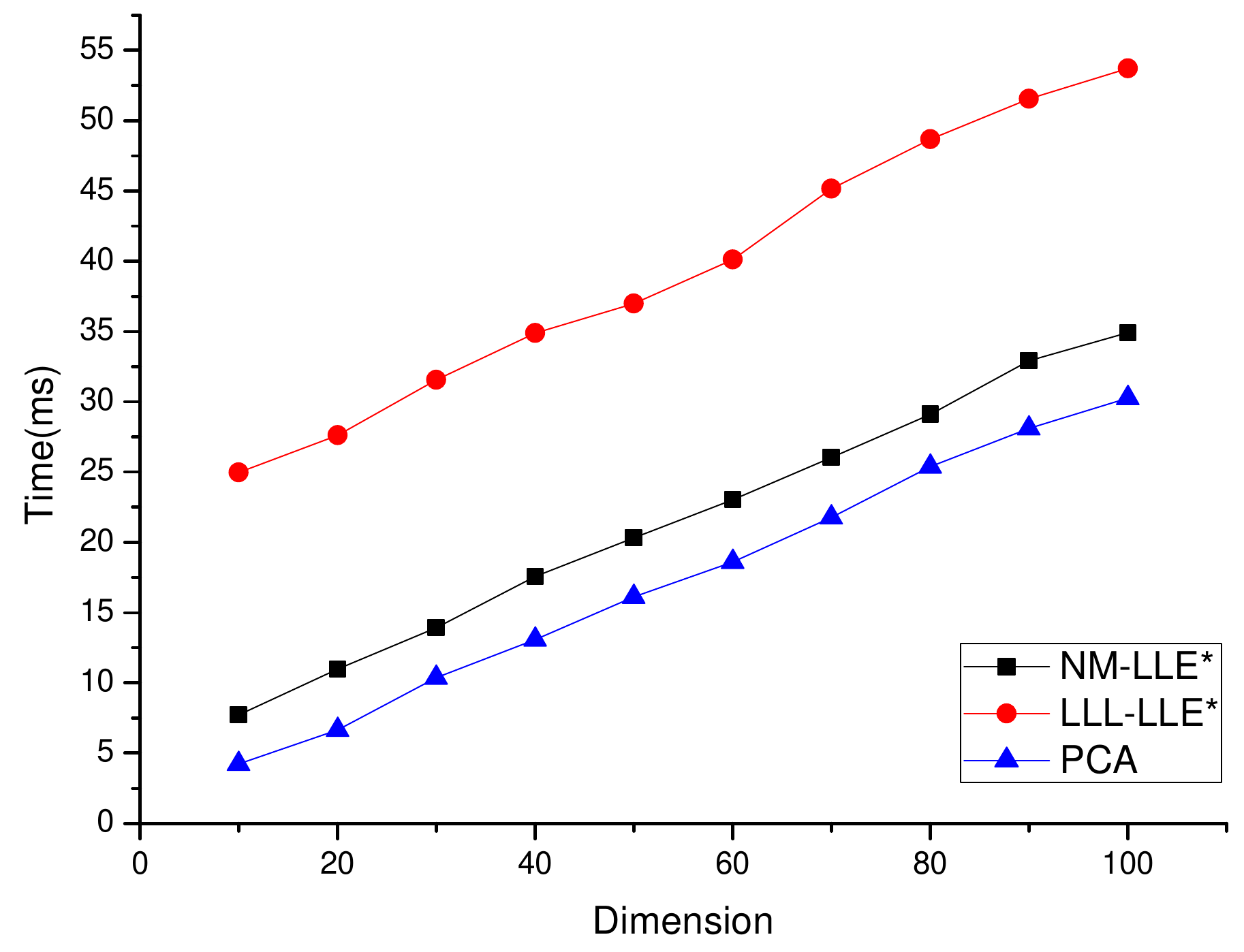}}
\label{fig3}
\caption{The time consuming of the four methods and original feature in Cifar-10}
\end{figure}

\begin{figure}[htbp]
\centerline{\includegraphics[width=3.2 in]{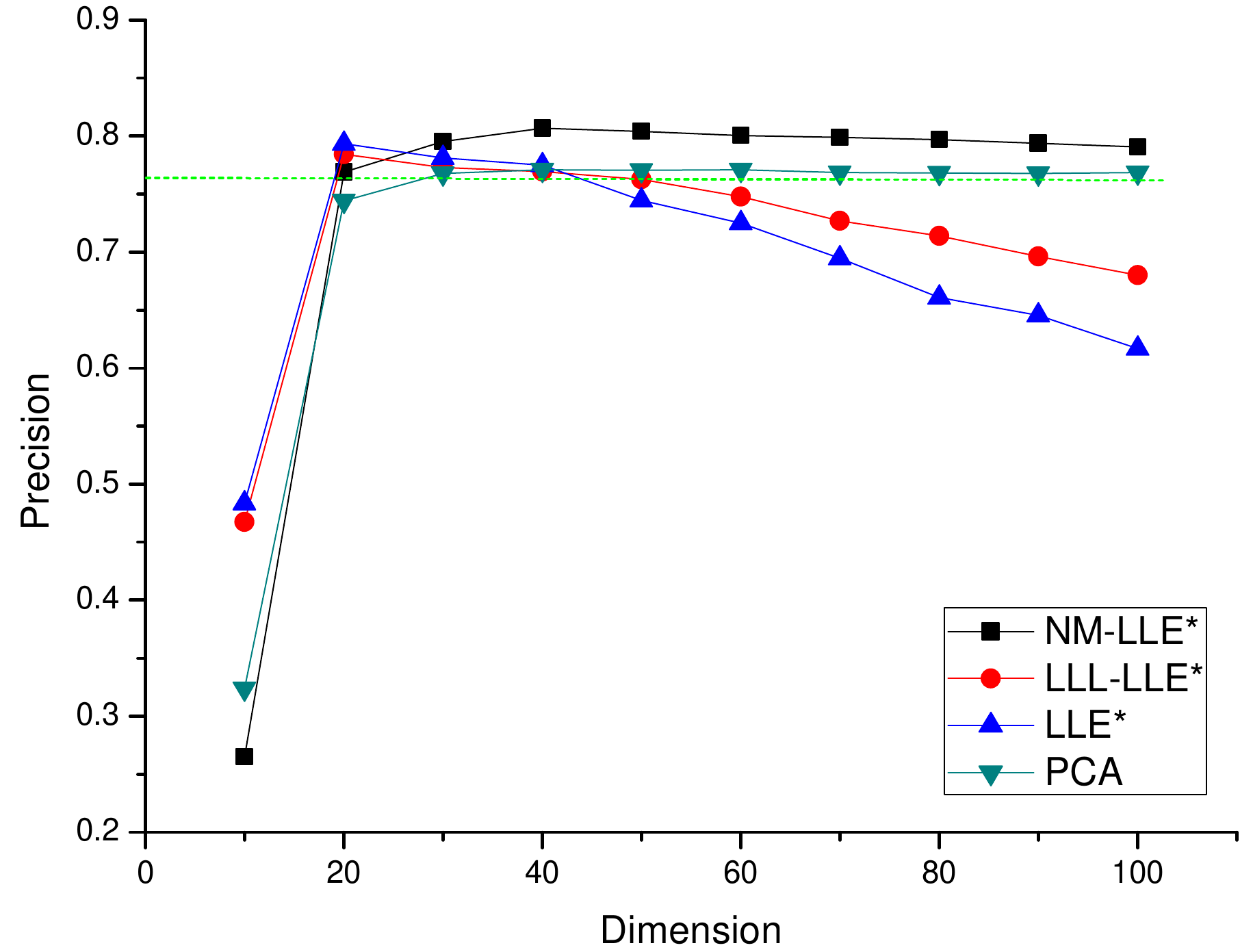}}
\label{fig3}
\caption{The precision of the four methods and original feature in Corel-1K}
\end{figure}

\begin{figure}[htbp]
\centerline{\includegraphics[width=3.2 in]{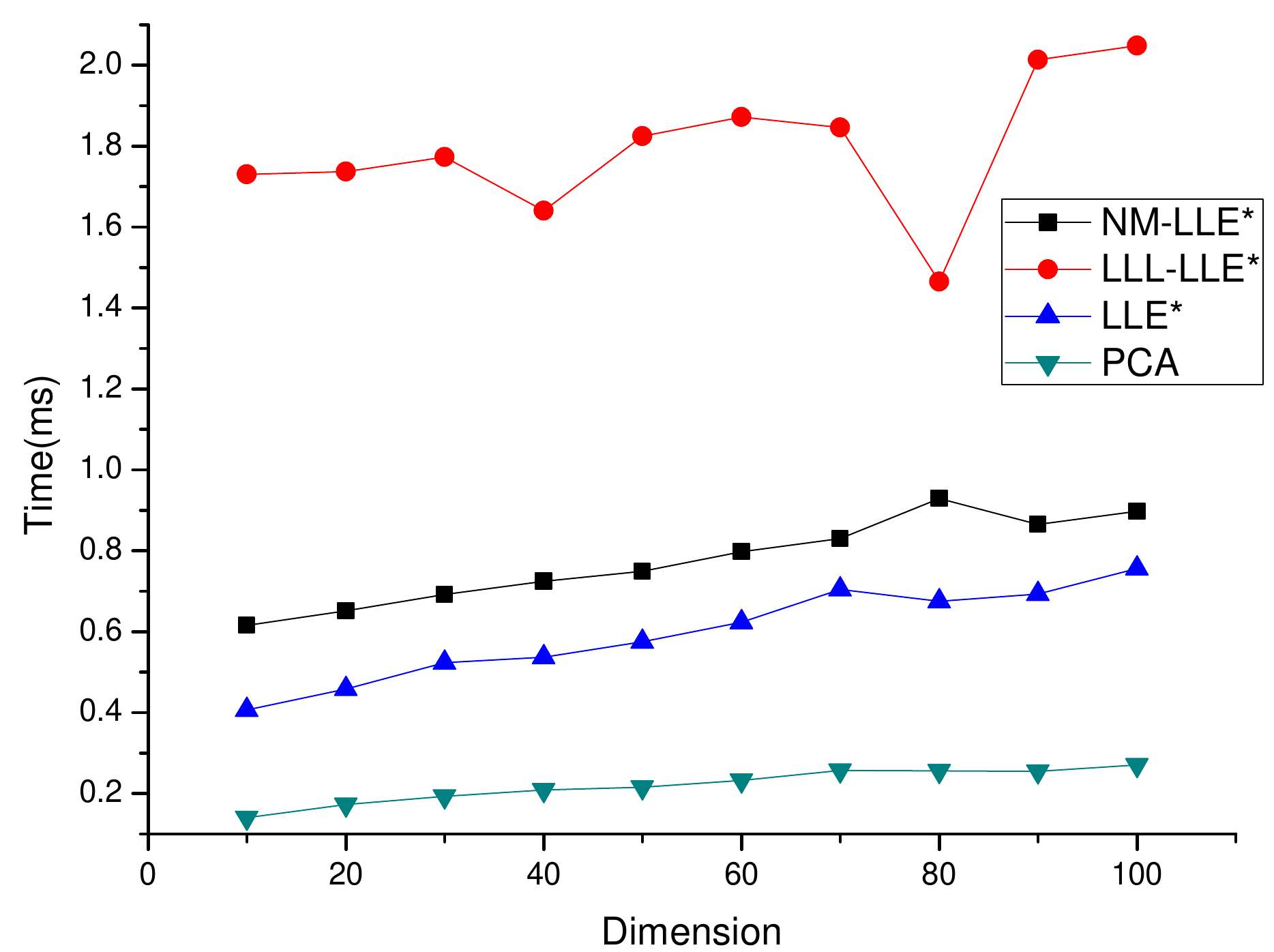}}
\label{fig3}
\caption{The time consuming of the four methods and original feature in Corel-1K}
\end{figure}

\begin{figure}[htbp]
\centerline{\includegraphics[width=3.2 in]{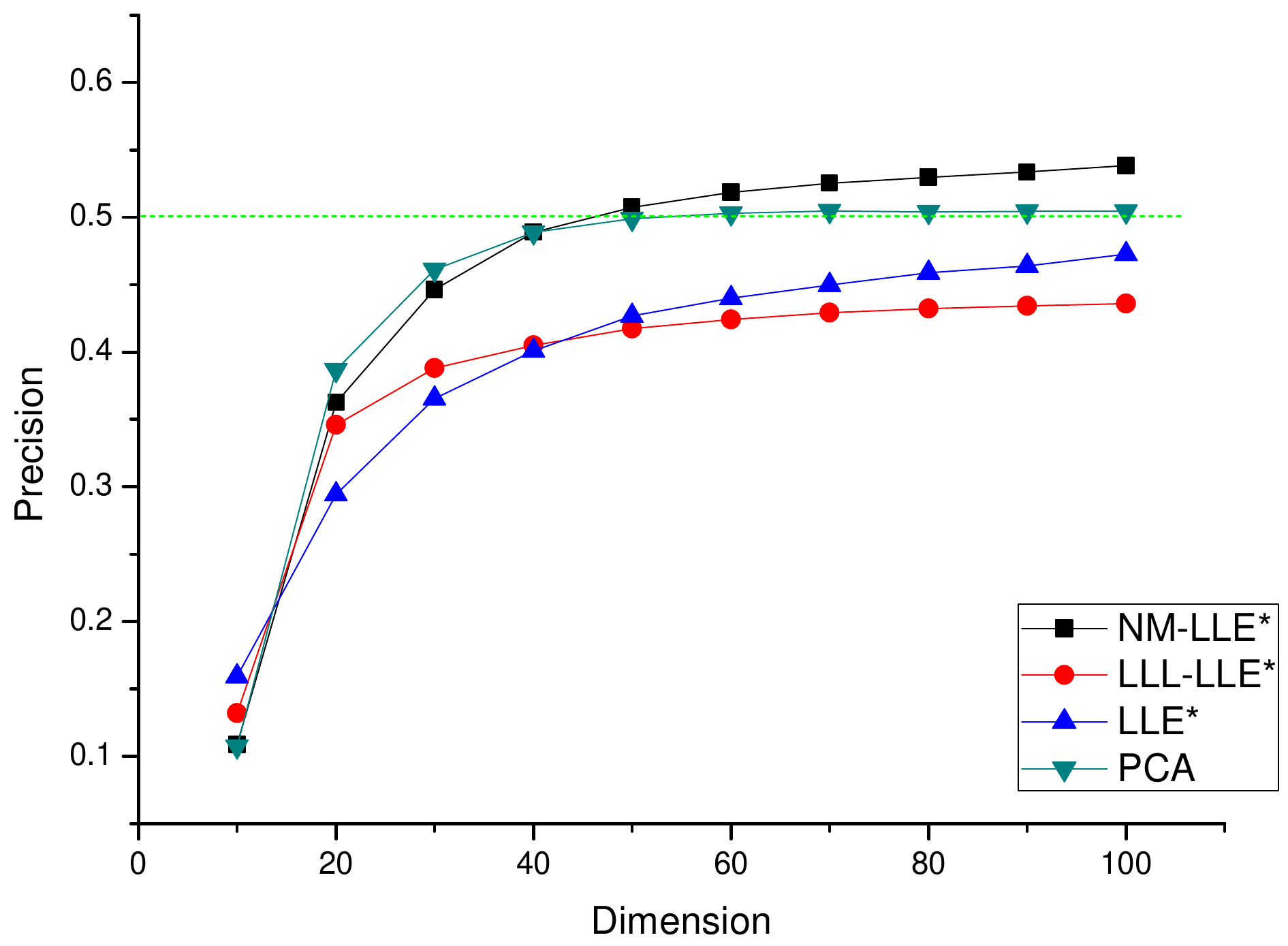}}
\label{fig3}
\caption{The precision of the four methods and original feature in Corel-10K}
\end{figure}

\begin{figure}[htbp]
\centerline{\includegraphics[width=3.2 in]{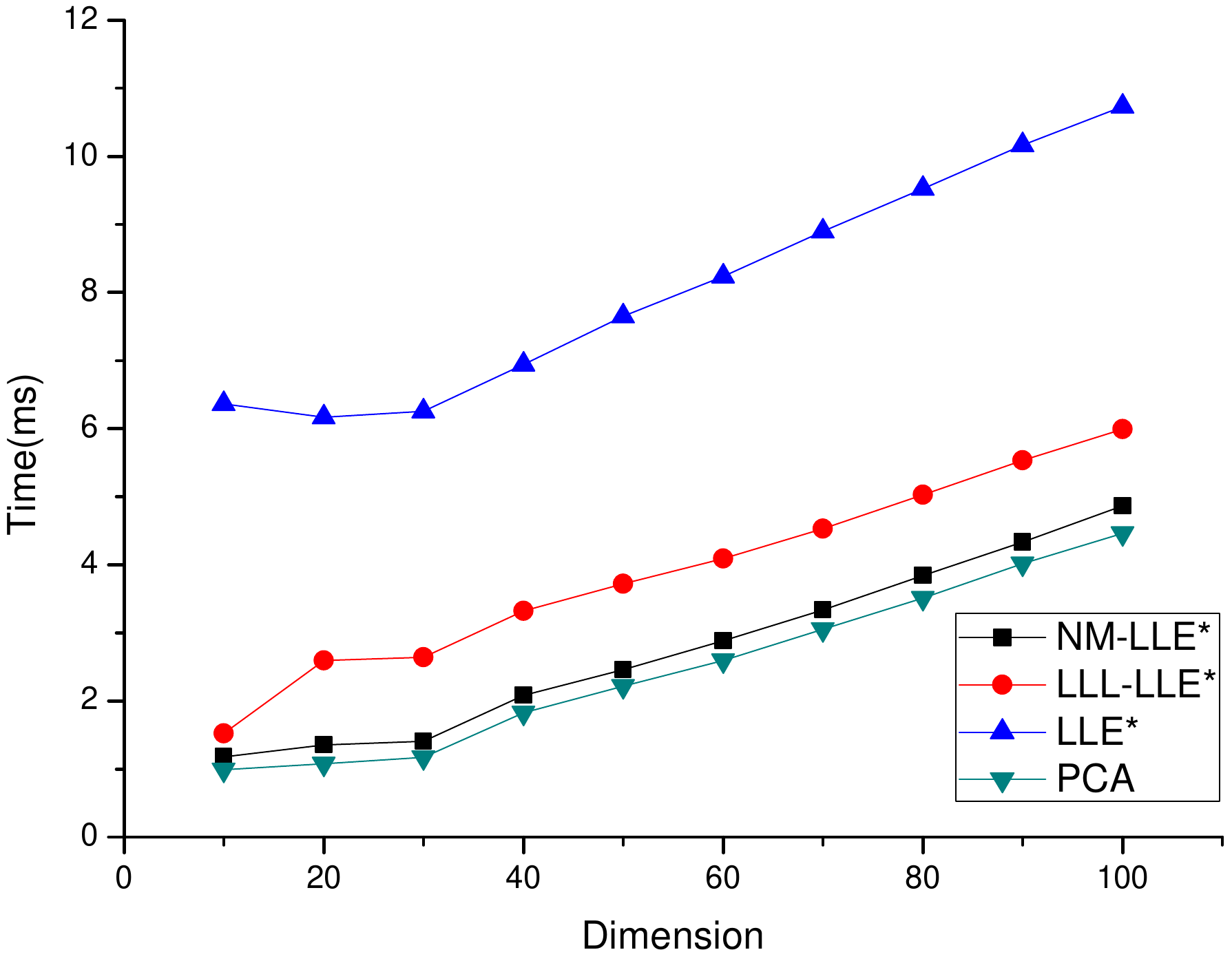}}
\label{fig3}
\caption{The time consuming of the four methods and original feature in Corel-10K}
\end{figure}

NM-LLE* always achieves highest Mea.P. and Max.P. among the four dimensionality reduction methods and original feature which are listed in Table \ref{tab2}. As shown in Table \ref{tab2} and Fig. 1, 3, 5. LLL method and LLE* can preserve manifold structure of images (features). However, NM is more suitable for retrieval and recognition applications. Fig. illustrate the precision of intrinsic dimensionality of data feature may get higher precision than that of other dimensions. The time consuming of different methods are list in Table \ref{tab3} and Fig. 2, 4, 6. As shown in Table \ref{tab3}, NM out-of-sample approach is acceptable for online retrieval. LLL, LLE* and original feature cannot be applied for online retrieval because of the expensive time consuming. Fig. show that the four methods of retrieval time are more while the dimensions of features are increased higher. The above analysis illustrates that our NM-LLE* is effective and efficient in image retrieval.

\begin{center}
\begin{table*}[htbp]
\caption{The time consuming of four dimensionality reduction methods in three datasets (ms) }
\centerline{
\begin{tabular}
{cccccc}
\hline
\multirow{2}{*}{Datasets}&
\multicolumn{4}{c}{Methods} &
\multirow{2}{*}{Original Feature} \\
\cline{2-5}
& NM-LLE*
& LLL-LLE*
& LLE*
& PCA
\\
\hline
Corel-1K&
0.7753&
1.7950&
0.5955&
0.2202&
115.2\\
Corel-10K&
2.7756&
3.8983&
8.0966&
2.4935&
694.8\\
Cifar-10&
21.66&
39.53&
780&
17.47&
1311.6\\
\hline
\end{tabular}}
\label{tab3}
\end{table*}
\end{center}

\section{Conclusion}
    Out-of-sample problem in nonlinear dimensionality reduction is an important research in machine learning. A good out-of-sample method should be efficient and effective. We first propose an adaptive locally linear embedding by curvature neighborhood. For out-of-sample problem of LLE*, we proposed a neural network based method (NM) in image retrieval task. Three benchmark datasets illustrate that NM+LLE* method can achieve higher precision than other state-of-art methods in both various image and sample size in datasets. However, we point out that our NM method cannot keep the manifold structure of real-word and hand-crafted data, which is not suitable for manifold learning problem. The above problem will be considered in our further work.

%\bibliographystyle{IEEEtran}


%\vfill


\end{document}